\documentclass[letterpaper, 10 pt, conference]{ieeeconf}  

\IEEEoverridecommandlockouts                              

\usepackage{graphics} 
\usepackage{epsfig} 
\usepackage{times} 
\usepackage{amsmath,amsfonts,amssymb,mathtools,leftidx}  
\usepackage{nccmath} 
\usepackage{mathrsfs}
\usepackage{xcolor}
\usepackage[ruled,vlined,linesnumbered]{algorithm2e}
\usepackage{cite}
\usepackage{url}
\usepackage{todonotes}
\usepackage{booktabs}
\usepackage[a-1b]{pdfx}
\usepackage{balance}

\title{\LARGE \bf
Koopman Operators for Modeling and Control of Soft Robotics
}

\author{Lu Shi, Zhichao Liu, and Konstantinos Karydis
\thanks{The authors are with the Dept. of Electrical and Computer Engineering, University of California, Riverside. 
	Email: {\{lshi024, zliu157\}@ucr.edu}, kkarydis@ece.ucr.edu. 
}}

\begin{document}

\maketitle
\thispagestyle{empty}
\pagestyle{empty}

\begin{abstract}
\textbf{Purpose of review}: We review recent advances in algorithmic development and validation for modeling and control of soft robots leveraging the Koopman operator theory.
 
\textbf{Recent findings}: We identify the following trends in recent research efforts in this area. (1) The design of lifting functions used in the data-driven approximation of the Koopman operator is critical for soft robots. (2) Robustness considerations are emphasized. Works are proposed to reduce the effect of uncertainty and noise during the process of modeling and control. (3) The Koopman operator has been embedded into different model-based control structures to drive the soft robots. 

\textbf{Summary:} 
Because of their compliance and nonlinearities, modeling and control of soft robots face key challenges. To resolve these challenges, Koopman operator-based approaches have been proposed, in an effort to express the nonlinear system in a linear manner. The Koopman operator enables global linearization to reduce nonlinearities and/or serves as model constraints in model-based control algorithms for soft robots. Various implementations in soft robotic systems are illustrated and summarized in the review.
\end{abstract}

\section{Introduction}\label{sec:intro}
Soft robots have been receiving significant attention over the years, and have grown into an important research topic.\
In contrast to their rigid counterparts, soft robots have bodies made out of intrinsically soft/or extensible materials, which exhibit unprecedented adaptation to complex environments and can absorb impact energy for safe interactions with the environment, other robots, and even humans~\cite{rus2015design}. Soft robots have been applied in all aspects of robotic research, including but not limited to assistive robotics~\cite{polygerinos2013towards, kokkoni2020development,mucchiani2022closed}, grasping~\cite{shintake2018soft, shi2022CASE, liu2022safely}, ground mobility~\cite{shepherd2011multigait, lu2018bioinspired}, legged locomotion~\cite{drotman2021electronics,liu2020sorx}, aerial robots~\cite{nguyen2022adopting} and underwater robots~\cite{li2021self}.

One of the prevailing challenges relates to their modeling and control~\cite{hughes2016soft}. To facilitate model-based control, researchers have proposed several quantitative modeling methods to parameterize soft robots (mostly continuum robots) using analytical models. For instance, piecewise constant curvature models~\cite{jones2006kinematics, webster2010design} with extensions to physical interactions~\cite{katzschmann2019dynamic, della2020model} have been successfully applied in modeling dynamic motion of soft robots. Further, numerical models have been utilized to discretize the geometry and achieve approximations. Cosserat rod theory~\cite{trivedi2008geometrically, rucker2010geometrically, giorelli2012two, janabi2021cosserat} and its extension to multiple degrees~\cite{black2017parallel}, discrete Cosserat models~\cite{renda2018discrete, armanini2021discrete}, polynomial curvature models~\cite{della2019control, della2020soft}, and rigid body approximations~\cite{roesthuis2016steering, nuelle2020modeling} have been proposed to model soft robots numerically. Finite element models have also been applied in modeling soft robots using industrial simulators~\cite{duriez2013control}, as well as reduced-order models~\cite{goury2018fast, katzschmann2019dynamically, tonkens2021soft}. However, the effectiveness of these methods relies on the accuracy of measurements of the actuator (e.g., length, strain, or curvature), if not relying on feedback~\cite{wang2021survey}.
Due to the infinite degrees of freedom of soft robots, accurate first-principles-based models are hard to derive. Moreover, in most cases, system nonlinearities pose additional difficulties to controller design~\cite{mucchiani2022closed}. One way to resolve such constraints is by considering data-driven methods~\cite{karydis2015IJRR, karydis2017ISER}.

Data-driven methods provide promising solutions to soft robot modeling, with the merits of alleviating requirements for sensory data while addressing sensor noise~\cite{wang2021survey}. Data-driven modeling methods for soft robots are generally categorized in two ways: 1) iterative kinematic mappings and 2) learning methods. The former utilizes iterative optimization to refine the desired mapping relationships among actuation, configuration and task spaces, such as iterative Jacobian matrix estimation~\cite{yip2014model, yip2016model} and adaptive Kalman filters~\cite{li2017model, shin2017adaptive, loo2019h, franco2021adaptive}. Learning methods for soft robots modeling include Feed-forward Neural Networks (FNN)~\cite{giorelli2013feed, giorelli2015neural}, Locally Weighted Projection Regression (LWPR)~\cite{lee2017nonparametric, ho2018localized} and Gaussian Process Regression (GPR)~\cite{fang2019vision}. Analytical models with learning-based extensions are also applied to improve the dynamic modeling of soft robots~\cite{braganza2007neural, queisser2014active, reinhart2016hybrid, tang2019novel}. In recent years, Reinforcement Learning (RL) has also been studied to model soft robots, such as model-based methods~\cite{polydoros2017survey, thuruthel2018model, wu2020position} and their model-free counterparts~\cite{you2017model, satheeshbabu2019open, satheeshbabu2020continuous, jiang2021hierarchical}.

Other than machine learning approaches, Koopman spectral theory~\cite{koopman1931koopman,mezic2005spectral,mezic2004comparison} has emerged as a dominant perspective over the past decade, in which nonlinear dynamics are represented in terms of an infinite-dimensional linear operator acting on the space of all possible measurement functions of the system~\cite{taylor202ALreview,brunton2022koopmanReview,bevanda2021koopmanReview,otto2021koopmanReview,kaiser2020operatorReview,brunton2019DDReview}. Recent research efforts have sought to use the Koopman operator in robotics. Examples include modeling and control of a tail-actuated robotic fish~\cite{mamakoukas2021Taylor}, trajectory control of micro-aerial vehicles~\cite{shi2020CCTA}, dynamics estimation for a spherical robot~\cite{abraham2017RKexample} as well as model extraction for a simulated lunar lander system~\cite{broad2020RKexample}. 

Recent advances in data-driven estimation approaches of the Koopman operator enable embedding the Koopman operator theory to soft robotic systems. First, instead of linearization at a certain point (local linearization), the Koopman operator is able to evolve a nonlinear system with full fidelity throughout the state space (\textit{global linearization}). This makes Koopman operators an attractive approach for obtaining linear representations of complex and/or nonlinear systems. By expressing a system's dynamics as a linear model, the stability properties can be more readily analyzed~\cite{mamakoukas2019local}, and various well-designed reliable controllers can be easily obtained~\cite{mauroy2016global}. Second, via mapping the original unknown system to the Koopman space and approximating the operator from measurements, the Koopman operator theory can be used as a \textit{data-driven model extraction} method. Both of the two aforementioned properties make the Koopman operator well-suited for implementation in soft robotics whereby complexity in modeling as well as control and nonlinearities are key challenges. 

The main goal of this review is to offer an overview of recent advances in implementing the Koopman operator in soft robotic systems. In Section~\ref{sec: background}, we introduce the theoretical foundations of the Koopman operator theory. In Section~\ref{sec: modelID}, we present the implementation details of using the Koopman operator to extract model descriptions of soft robots. In Section~\ref{sec: controller}, different types of Koopman-based control structures and examples are introduced. Finally, in Section~\ref{sec: conclusion}, we propose several possible future directions and discuss the challenges that can be addressed when applying the Koopman operator in the modeling and control of soft robotic systems.

\section{Background}\label{sec: background}
\begin{figure}
    \centering
    \includegraphics[width = 0.48\textwidth]{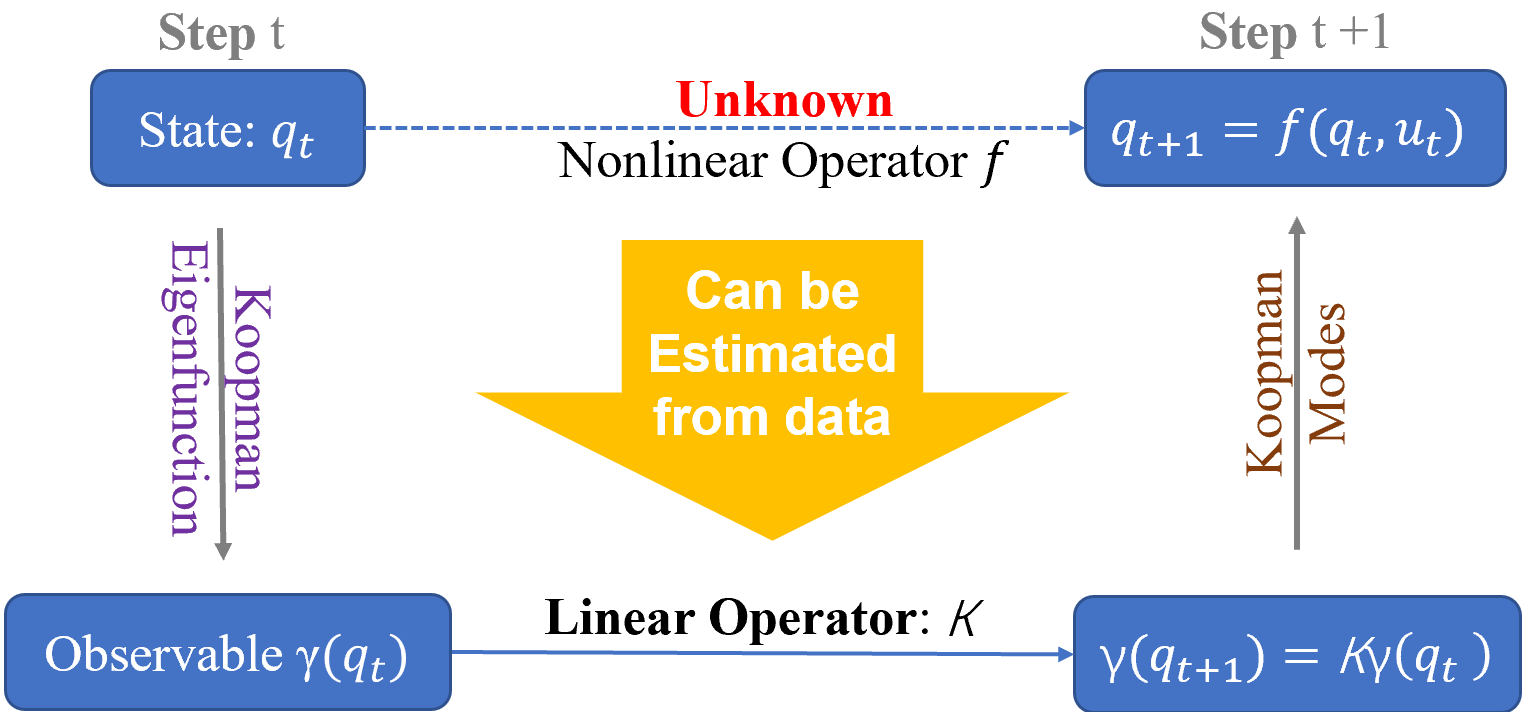}
    \caption{Overview of the Koopman operator theory.}
    \label{fig:Koopman}
\end{figure}

In this section, we give an overview of key relevant results on the Koopman operator as shown in Fig.~\ref{fig:Koopman}, the Koopman generator and related data-driven approximation approaches.

\subsection{Koopman Operator}\label{subsec:Koopman}
Consider a dynamical system with state vector $q \in \mathbb{R}^{n_{q}}$,
\begin{equation}\label{eq:system}
    q_{t+1} = f(q_t)\enspace \text{or}\enspace  \frac{d}{dt} q = f(q)\enspace, 
\end{equation}
for the discrete-time and continuous-time settings, respectively. Define a set of observables $\gamma \in \mathcal{F}$ to be the \textit{mapped} or \textit{lifted} states. The evolution of observables $\gamma$ using the infinite-dimensional Koopman operator $\mathcal{K}:\mathcal{F} \to \mathcal{F}$ can be written as $$\mathcal{K} \gamma(q_t)= \gamma(f(q_t))\enspace \text{or} \enspace \frac{d}{dt} \gamma(q) = \mathcal{K}\gamma(q)\enspace.$$

If we further define the lifted states $\gamma(q)$ as \textit{new} states $z = \gamma(q)$ in this linear Koopman space, the original nonlinear problem is transferred into a linear problem in this new space and various traditional linear controllers can be designed and implemented. 

The Koopman operator can also be utilized to identify the dynamical system of the nonlinear form. Under the full-observability assumption that there exists one of the observables $\gamma(q)=q$, $N$ Koopman-based items including Koopman modes $v_n$, eigenvalues $\lambda _n$ and eigenfunctions $\varphi_n$ can be obtained by decomposing the Koopman operator~\cite{williams2015EDMD}. Finally, the unknown nonlinear operator $f$ in \eqref{eq:system} can be approximated as the combination of the Koopman-based items as 
\begin{equation}\label{eq:extendPredicion}
\begin{medsize}
    q_{t+1} = \gamma(f(q_t))= \mathcal{K} \gamma(q_t) \to q_{t+1}  =\sum_{n=1}^N v_n\lambda_n\varphi_n(q_t)\enspace . 
\end{medsize}
\end{equation}

\subsection{Extended Mode Decomposition (EDMD)}\label{subsec:EDMD}
In practice, we typically have access to the discretely sampled measurement data of the system, which are used to obtain a finite-dimensional approximated matrix of the infinite-dimensional Koopman operator. Although approximating the Koopman operator induces errors in the system propagation,
it has been shown that the linear model is able to evolve the original system with acceptable accuracy~\cite{klus2016convergence,korda2018convergence}. 

Of all the approximation approaches, one of the most popular algorithms is the Extended Dynamic Mode Decomposition (EDMD) approach~\cite{williams2015EDMD}. It describes how to learn those Koopman-based items in~\eqref{eq:extendPredicion} from data. Given state history {$q = [q_1,q_2,\dots,q_{M},q_{M+1}]$} (commonly referred to as ``snapshots"), $\mathcal{K}$ can be expressed as a finite-dimensional approximation $K:\mathcal{F}_N \to \mathcal{F}_N$ of the Koopman operator $\mathcal{K}:\mathcal{F} \to \mathcal{F}$ via EDMD. To do so, we need a dictionary of functions that lift state variables to the higher-dimensional space where observable dynamics is approximately linear. Define the dictionary $\mathbf{\Psi}(q_m)=[\psi_{1}(q_m), \ldots, \psi_{N}(q_m)]$, the Koopman operator can be approximated by minimizing the total residual between snapshots, 
$$J =\frac{1}{2} \sum_{m=1}^{M}\left(\mathbf{\Psi}\left(q_{m+1}\right)-\mathbf{\Psi}\left(q_{m}\right) K\right)^{2}\enspace.$$ 
This can be solved by truncated singular value decomposition, yielding
    \begin{equation}\label{eq:estimation}
        K \triangleq \boldsymbol{G}^{\dagger} \boldsymbol{A}, \enspace
        \text{where }
        \begin{cases}
      \boldsymbol{G}=\frac{1}{M} \sum_{m=1}^{M}  \mathbf{\Psi}_m^{*} \mathbf{\Psi}_m\enspace,\\
      \boldsymbol{A}=\frac{1}{M} \sum_{m=1}^{M}  \mathbf{\Psi}_m^{*}  \mathbf{\Psi}_{m+1}\enspace,
    \end{cases}
    \end{equation}
with $^\dagger$, $T$ and $^*$ denoting pseudoinverse, transpose and conjugate transpose, respectively. 
Obtained $K$ via~\eqref{eq:estimation}, we get 
\begin{equation}\label{eq:KoopmanDecomposition}
    \begin{cases}
        v_n = (w_n^*B)^T \enspace,\\
        \lambda_n \eta_n = K\eta_n \enspace,\\
        \varphi_n = \mathbf{\Psi}_t\eta_n \enspace,
    \end{cases}
\end{equation}
where $\eta_n$ is the $n$-th eigenvector, $w_n$ is the $n$-th left eigenvector of $K$ scaled so $w_n^T\eta_n = 1$, and $B$ is a weight matrix such that $q=(\mathbf{\Psi} B)^T$~\cite{williams2015EDMD}. 

The evolution of the original nonlinear system using the estimated Koopman operator is described by replacing expressions~\eqref{eq:KoopmanDecomposition} into~\eqref{eq:extendPredicion}. Control inputs can be readily incorporated into the definition of $\mathbf{\Psi}$ as an augmented state~\cite{proctor2018EDMDc}. 

If the lifting dictionary is chosen as $$\mathbf{\Psi}(q_m)=[e_1^Tq, \ldots, e_{n_{q}}^Tq_m,]$$ where $e_i$ is the $i-th$ unit vector in $\mathbb{R}^{n_{q}}$, the estimated Koopman operator generated by the previous procedure is same as that obtained by Dynamic Mode Decomposition (DMD)~\cite{williams2015EDMD}. Theoretically, DMD tries to get a linear approximation of the original system, i.e. $q_{t+1} = Kq_t$. The estimate operator $K$ is selected as the best-fit solution for
all available data. Similarly, the DMD-based method can be extended for inputs~\cite{proctor2016DMDc}.
\subsection{Koopman Generator}
To better understand the beauty of the theory, we will give a short introduction to its associated infinitesimal generator - the Koopman generator, which is more close to the idea when the Koopman operator was first proposed in fluid dynamics theory~\cite{koopman1931koopman}. 

Consider the system defined in \eqref{eq:system}, if the initial condition is defined as $q_0$, the solution to the ordinary differential equation at time $t$ can be denoted as $\phi^t(q_0)$. It is also referred to as the flow map of the system. Define the Koopman semigroup of operators $\{\mathcal{K}^t\}$ as $(\mathcal{K}^tg)(q) = g(\phi^t(q))$, the infinitesimal generator $\mathcal{L} \in \mathcal{F}$ of the semigroup is given by 
\begin{equation}\label{eq:generator}
    \mathcal{L}g = \lim_{t\to 0}\frac{1}{t}(\mathcal{K}^tg-g) = \frac{d}{dt}g = \sum_{i=1}^{N_q}\gamma_i\frac{\partial g}{\partial q_i} \enspace, 
\end{equation} 
it can be lifted to an infinite-dimensional function space $\mathcal{F}$, which is thus also the space of observables. 

The Koopman generator can be obtained by the reformulated EDMD - gEDMD~\cite{klus2020generator}. Similarly, a dictionary of lifting functions is defined as $\mathbf{\Psi}(q)$ while we define additionally according to \eqref{eq:generator}
that:
$$\dot{\mathbf{\Psi}}(q) = (\mathcal{L}\mathbf{\Psi})(q) = \sum_{i=1}^{N_q}\gamma_i(q)\frac{\partial \mathbf{\Psi}(q)}{\partial q_i}\enspace.$$
Collecting the matrices of $\mathbf{\Psi}(q)$ and its derivatives $\dot{\mathbf{\Psi}}(q)$ of all the measurements, the generator is estimated by solving a similar least-square problem as what we do when calculating the Koopman operator. 

\textbf{Comparison of the Koopman operator and its generator}: The data-driven extraction approach of the Koopman operator (EDMD) is originally proposed and propagated for the discrete-time system. It can be extended to the continuous-time operator via $\mathcal{K}_c = \log(\mathcal{K})/\Delta t$, where $\Delta t$ is the sampling time of the measurements. On the other hand, the Koopman generator is analyzed in a continuous manner. The benefit of using it is that the Koopman generator might be sparse even when the operator is not. However, the data-driven approximation of the Koopman generator requires analytical precomputation of the partial derivatives, which might introduce more prediction errors. The trade-off needs to be considered when selecting between the two versions of the Koopman theory.

\begin{table*}[h]
\vspace{0pt}
    \caption{Summary of recent advances in Koopman operators in soft robotics.}\label{tab:summary}
    \vspace{-6pt}
    \centering
    \begin{tabular}{c c c c}
        \toprule
        Paper  & Lifting Functions & Controller & Implementation Platform\\ 
        \midrule
       ~\cite{komeno2022deep}  & Neural Network & MPC & Soft Inverted Pendulum 
       \\
       \midrule
       ~\cite{castano2020swimmer} & Mechanics-inspired &--- & Soft Swimmer 
       \\
       \midrule    ~\cite{bruder2020TRO},~\cite{bruder2019arxiv},~\cite{bruder2019ICRA} & Monomials &MPC & Soft arm with laser pointer
       \\
       \midrule
       ~\cite{han2021desko} & Neural Network & MPC& Soft continuum 
       \\
       \midrule
       ~\cite{bruder2021loading} & Monomials & MPC & Granular Jamming Gripper 
       \\
       \midrule 
       ~\cite{chen2022offset} & Monomials & MPC & Soft Manipulator 
       \\
       \midrule
       ~\cite{haggerty2020arxiv} & Monomials & LQR& Soft Arm 
       \\ 
       \midrule
       \cite{shi2021ACDEDMD} &Mechanics-inspired &--- & Soft Leg
       \\
       \midrule
       \cite{shi2022CASE} &Mechanics-inspired & MPC & Soft Gripper 
\\
        \bottomrule
    \end{tabular}
\end{table*}

\section{Model Identification with Koopman Operator}\label{sec: modelID}
Soft robots are composed of compliant materials, instead of rigid links and joints. The compliance provides soft robots flexibility, softness as well as safety when working in close contact with environments and humans. But these properties also make it hard to obtain model representations for analysis and control. Approaches to obtain models of robots can be separated into two categories: physics-based methods (i.e. derived from first principles) and data-driven methods. The model construction from physics-based approaches can be complicated for soft robots given their infinite degrees of freedom and nonlinear behaviors. One way to solve the problem is to embed the Koopman operator theory into model extraction. A primary benefit of the Koopman-based techniques is that a description of an input-output relationship can be obtained from data without explicitly defining a system state. This is especially useful for obtaining reduced-order models of soft robots that have essentially infinite-dimensional kinematics. Various works have been implemented in this area including the modeling and control of soft grippers, soft arms, soft legs, soft swimmers, etc. A detailed description is listed in Tab.~\ref{tab:summary}. In this section, we emphasize the essential parts that need to be considered for explicit implementation and safety considerations in the practical examples. 

\subsection{Design of Lifting Functions}
A critical challenge inherent to all methods employing Koopman operator theory is the choice of a proper set of lifting functions (typically called the dictionary). The lifting functions are crucial because they serve as the basis to construct an infinite-dimensional linear approximation of the original (often nonlinear) system's state evolution. Poor choice of the lifting functions can significantly impact the estimation of the Koopman operator and hence the accuracy of higher-dimensional linearized dynamics.

Various distinctive approaches have been tested and implemented in different soft robotic systems to identify their dynamics with Koopman-based methods. Existing works regarding the construction of dictionaries fall under the following main directions.

\begin{enumerate}
    \item \textbf{Empirical Selection:} The most frequently-used lifting functions in Koopman-based soft-robot modeling applications are the monomials~\cite{bruder2020TRO,bruder2019ICRA,kamenar2020five,chen2022offset,haggerty2020arxiv} and polynomials~\cite{bruder2021loading}. 
    The maximum degree and dimensions are selected manually based on the properties of the underlying system. Fewer degrees might lead to insufficient and inaccurate estimation while more degrees can cause higher computation complexity and more possibility of overfitting. Thus, the dimension needs to be determined empirically or by comparing multiple candidates~\cite{kamenar2020five}.

    \item \textbf{Mechanics-inspired Selection:}
    Inspired by various physics-based approaches, it is plausible to consider that the dynamic of the soft robot has the same structure as that of an approximate averaged dynamical model of a similar rigid robot. For example, it has been shown possible that a soft swimmer has a similar model to a rigid robotic fish~\cite{castano2020swimmer}. Although it is known that the dynamics differ between the soft and the rigid systems, the overall model structure is sufficient enough to design the basis functions used to estimate observables. A similar rigid robot may not always be accessible, yet, robotic systems have certain characteristic properties, e.g., degrees of freedom, configuration spaces, and workspaces, that can be acquired without knowledge of their exact dynamical models. These properties reveal fundamental information about system states and dynamics and can provide intuition on how to select lifting functions required for Koopman operator approximation. In~\cite{shi2021ACDEDMD}, the authors have shown how fundamental topological spaces and Cartesian products thereof can be mapped to a basis of Hermite polynomials and Kronecker products thereof which serve as the dictionary of lifting functions.

    \item \textbf{Machine Learning Approaches:}
    Another direction is to rely on machine learning methods to derive the dictionary~\cite{han2021desko} or the eigenfunctions directly~\cite{komeno2022deep}. The vector $\mathbf{\Psi}$ in Sec.~\ref{sec: background} is defined as a neural network, and parameters are tuned with offline training data to achieve high prediction accuracy. While such methods can have a stronger generalization capacity, they require significant tuning (e.g., in the case of neural networks the number of layers, number of units per layer, etc.), and large amounts of training data. The latter might in practice pose a challenge in robotics applications where data are in principle small. 

\end{enumerate}

\textbf{Summary:} Multiple algorithms are proposed to find the dictionary of lifting functions. Each of them has its own benefits and disadvantages to be utilized and the selection should be determined based on the system and measurements. If the offline dataset is large as well as diverse enough, in other words, the offline training data cover as various cases as possible, the Machine-Learning-based dictionary could achieve high accuracy with sufficient time. If a similar rigid robot is available or the workspace of the system is known, the Mechanics-inspired approaches should be the best choice. Otherwise, the lifting functions can be designed based on experience or selected from multiple candidate dictionaries.

\subsection{Robustness Considerations}
Deriving guarantees and robustness for methods that rely on extracting dynamics from data are important as the efficiency and quality of measurements will have a significant impact on the performance of the estimated model and hence the controller design.  

Different considerations are proposed to emphasize the robustness when utilizing Koopman-based approaches for model extractions. In~\cite{han2021desko}, a deep stochastic Koopman operator is introduced to guarantee the stability of nonlinear stochastic systems. A probabilistic neural network is designed to estimate the distribution of the observables and the distribution is propagated by the Koopman matrices. The parameters of the probabilistic neural network as well as the Koopman operator are optimized during the training process.

Another design~\cite{chen2022offset} to minimize the effect of modeling uncertainties and external disturbance is to include the usage of the Kalman filter. First, an augmented model is derived to include the disturbance model in the ordinary Koopman structure, and then a Kalman filter is adapted to estimate the observables and disturbances simultaneously.

Sparse models can be acquired from data driven model extraction approaches for soft robots. Sparsity is desirable in terms of space complexity, which can enable the usage of a higher-dimensional dictionary. However, it can lead to an ill-conditioned matrix and even singularities. An ill-conditioned operator is sensitive to every measurement, which can carry a cost in prediction accuracy especially when the measurements are noisy. One way to deal with this problem is to add an optional projection after each approximation step to make sure the models learned are close to the lifted space~\cite{bruder2020TRO}. 

There also exist some works to derive guarantees for methods employing the Koopman operator, including investigation of convergence of estimation~\cite{korda2018convergence,korda2018KoopmanMPC,peitz2019switchsystem} and global error bounds for the operator~\cite{mamakoukas2021Taylor,mamakoukas2020stable}. Another direction is investigating the 
prediction error of, or providing robustness guarantees for, the perturbed systems' performance when the data used for modeling are noisy~\cite{shi2021noise}. However, such aforementioned investigations remain to be implemented and tested in soft robotic systems. 

\begin{figure}
    \centering
    \includegraphics[width = 0.49\textwidth]{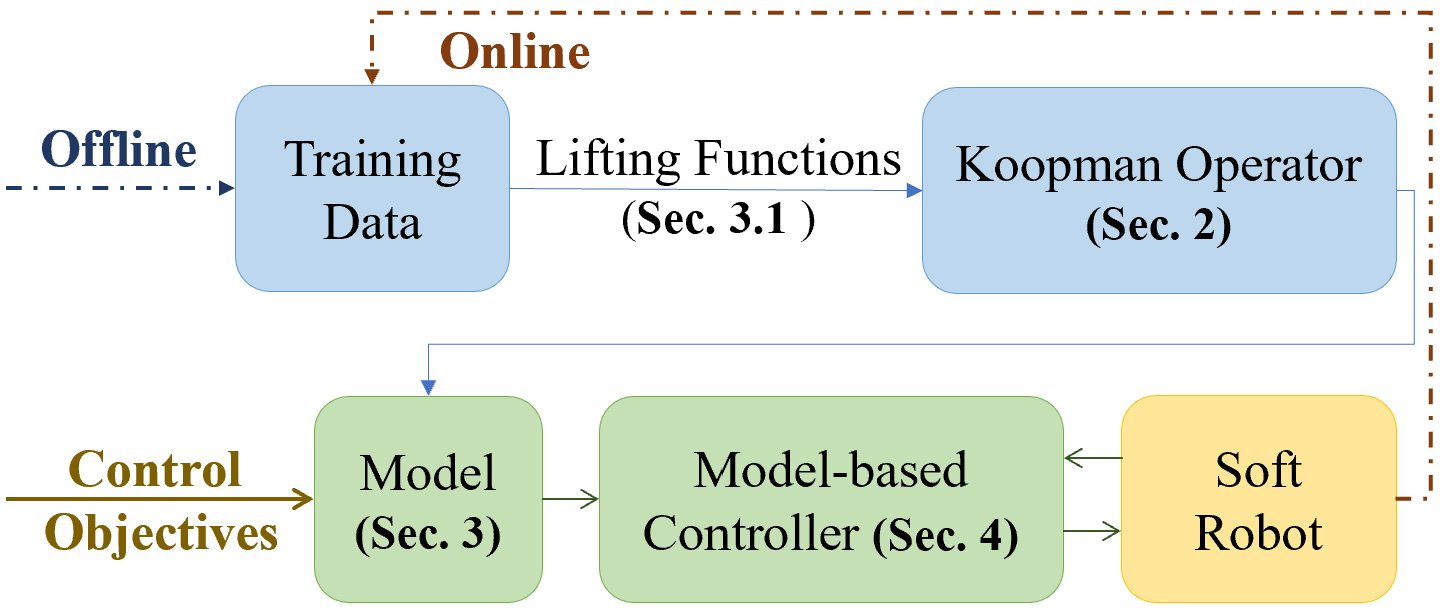}
    \caption{Overall procedure of modeling and control of soft robotic systems with the Koopman operator theory. The dashed lines indicate that the training data could be acquired either online or offline. Also, note that the model learned by the Koopman operator can be either linear or nonlinear determined by the control objective.}    \label{fig:overall_procedure}
\end{figure}

\section{Controller Design} \label{sec: controller}
Soft robots differ from rigid-bodies in several ways that make them uniquely difficult to control. Soft robots exhibit nonlinear characteristics that are negligible in rigid materials, and they span such a wide range of designs which poses great challenges to the generalization ability of the designed controller. These differences render many of the standard approaches used to control rigid-bodied robots insufficient for soft robots.

A system model enables the design of model-based controllers that leverage model predictions to choose suitable control inputs for a given task. As introduced in Sec.~\ref{sec: modelID}, the Koopman operator can obtain either a linear model or a nonlinear model from data, which can be easily embedded into the design of corresponding control structures. The Koopman operator is naturally formed for unforced systems. To extend it to forced cases for soft robots, the inputs can be either augmented to state and be lifted alongside the state~\cite{komeno2022deep,han2021desko,shi2021ACDEDMD,shi2022CASE} or be appended but not lifted to ensure that they appear linearly in the resulting model~\cite{bruder2020TRO,bruder2019ICRA,bruder2019arxiv,haggerty2020arxiv,chen2022offset}. In this section, we give an overview of notable options for control-oriented frameworks based on the Koopman operator theory.

\subsection{MPC} In particular, model-based controllers can anticipate future events, allowing them to optimally choose control inputs over a finite time horizon. A popular model-based control design technique that illustrates the beauty of prediction is model predictive control (MPC), wherein one optimizes the control input over a finite time horizon, applies that input for a single timestep and repeats the optimization process. The key component of, and the main item that affects the performance of, MPC is the ``model." The Koopman operator, which takes into account new measurements and/or offline measurements, can serve as the model constraints in the MPC structure formulation.

Incorporating Koopman-based
models into MPC design has been first introduced in~\cite{korda2018KoopmanMPC}. The Koopman-based MPC is widely used in soft platforms for either linear MPC~\cite{bruder2020TRO,bruder2021loading,chen2022offset, bruder2019arxiv} or nonlinear MPC~\cite{bruder2020TRO,shi2022CASE}. In~\cite{komeno2022deep}, a soft inverted pendulum is controlled to be suspended stably in the inverted position. The Koopman-based MPC is also implemented in the soft continuum robots, the soft gripper to track desired trajectories~\cite{han2021desko, bruder2019arxiv,bruder2019ICRA, chen2022offset,shi2022CASE}. In these cases, the cost functions are always defined to minimize the difference between the predicted states obtained using the Koopman operator and desired trajectories or a target position.



\subsection{LQR} 
The linear model learned from data is appropriate to be implemented in optimal control methods like the Linear Quadratic Regulator (LQR)~\cite{anderson2007LQR}. The LQR algorithm is essentially an automated way of finding an appropriate state-feedback controller, which reduces the amount of work done by the control systems engineer to optimize the controller. In~\cite{haggerty2020arxiv}, a reduced-order linear model of a helically actuated, inertial soft
arm is obtained by the Koopman operator and serves as the model constraints to solve the discrete-time algebraic Riccati equation. The resulting control signal drives the soft arm to achieve two desired states. However, the scenario is limited to offline training and open-loop control where the input sequences are determined in simulation and deployed to the robot. An online-updated design for the soft robot and closed-loop optimal control theory with the Koopman operator remains an open direction. 

\textbf{Summary:} The Koopman operator has been used for model-based control of dynamical systems, including feedback stabilization~\cite{huang2019feedback}, optimal control~\cite{abraham2019optimalControl,abraham2017RKexample}, model predictive control~\cite{arbabi2018mpc,korda2018KoopmanMPC} and hierarchical adaptive control~\cite{shi2020CCTA}, etc. These works illustrate the benefits of model-based control with the Koopman operator theory while, unfortunately, those control algorithms are not fully tested in soft platforms, which indicates directions of future work in this area.

\section{Discussion and Conclusion}\label{sec: conclusion}
The Koopman operator theory has advanced considerably in recent years. In this review, we have explored its usage to represent nonlinear dynamical systems in a linear framework. Several data-driven approaches to estimate the Koopman operator are introduced. We have also summarized the detailed implementation process of the Koopman framework in soft robots and practical considerations. Finally, various control architectures that deploy the soft robotic systems to achieve different control objectives with the Koopman operator theory are illustrated with specific examples.

Despite the incredible promise of the Koopman operator, its application in soft robotic systems is still a very open field under active exploration. As concluded in Tab.~\ref{tab:summary}, current applications are limited to soft robots with relatively simple structures. Investigations of and implementation on more complicated systems are promising directions to be investigated in the future. The extended platform can include rigid robots with soft parts and even complete elaborate compliant robots. On the other hand, the design of the Koopman-based controllers can be extended to more than MPC and optimal control. Various linear or nonlinear control architectures have the potential to be embedded into the soft robots with the Koopman operator theory. 

Along with those more general open questions, there are several problems in terms of detailed implementation in the soft platforms that can be emphasized. First, the compliance of the material makes soft robots usually react slower than the rigid body, which poses a challenge to the time efficiency of the learned model and controller. In other words, if the sampling frequency is too large, it would be hard for the learned model to capture the actual relation between inputs and states as the states will be collected before the effect of inputs reflects in the output states. If the sampling rate is too small, the controller performance will be weakened. A proper and general way to find the optimal sampling rate is required and worth to be investigated. In addition, the accuracy of measurements is significant for all the data-driven approaches. However, the sensors and observers for soft robots are limited. Other than different types of pressure sensors, researchers tend to use motion capture systems or cameras to obtain position (and orientation) information of tips or selected points of the soft robot. But the velocity and/or higher-order derivatives are hard to be obtained accurately and directly from sensors, which currently makes it difficult to achieve aggressive control objectives for soft robots with the data-driven Koopman operator methods.

The applications of the Koopman operator theory in the modeling and control of soft robots are still an open and active research direction. More works and investigations are expected in the future. We hope that this review paper can serve as a stepping stone to inspire more works in the area.

\balance
\bibliographystyle{IEEEtran}
\bibliography{IEEEabrv,IEEEexample}

\end{document}